\documentclass[runningheads]{llncs}
\usepackage{comment}
\usepackage{graphicx}
\usepackage{amsmath}
\usepackage{amsfonts}
\usepackage{subcaption}
\usepackage{epstopdf}
\usepackage[table,xcdraw]{xcolor}
\usepackage{multirow}
\usepackage[pagebackref=true,breaklinks=true,letterpaper=true,colorlinks,bookmarks=false]{hyperref}
\usepackage{booktabs}

\definecolor{c2}{HTML}{FBD9BD}
\definecolor{c3}{HTML}{fe793d}
\definecolor{c4}{HTML}{eedeb0}
\definecolor{rouse}{rgb}{0.981,0.961,0.941}

\makeatletter
\newcommand*\bigcdot{\mathpalette\bigcdot@{.5}}
\newcommand*\bigcdot@[2]{\mathbin{\vcenter{\hbox{\scalebox{#2}{$\m@th#1\bullet$}}}}}
\makeatother

\begin{document}
\title{Concealed Object Segmentation with Hierarchical Coherence Modeling
% \thanks{This work was supported by the National Science Foundation of China under grant 62201538 and Natural Science Foundation of Shandong Province under grant ZR2022QF006.}
}

% INITIAL SUBMISSION 
% \def\CICAISubNumber{19}  % Insert your submission number here
% %\begin{comment}
% \titlerunning{CICAI2022 submission ID \CICAISubNumber} 
% \authorrunning{CICAI2022 submission ID \CICAISubNumber} 
% \author{Anonymous CICAI submission}
% \institute{Paper ID \CICAISubNumber}
% %\end{comment}
%******************

% CAMERA READY SUBMISSION
% \begin{comment}
% \titlerunning{Abbreviated paper title}
% If the paper title is too long for the running head, you can set
% an abbreviated paper title here
%
\author{Fengyang Xiao$^{1,}$\thanks{First Author, $\dagger$ Corresponding Author}\orcidID{0009-0001-5770-8953} \and
Pan Zhang\inst{1}\orcidID{0009-0002-0721-4063} \and
Chunming He$^{2,\dagger}$\orcidID{0000-0001-6479-7109}\and
Runze Hu$^{3}$\orcidID{0000-0002-6366-3763} \and
Yutao Liu\inst{4}\orcidID{0000-0002-3066-1884}}
\authorrunning{F. Xiao et al.}
% First names are abbreviated in the running head.
% If there are more than two authors, 'et al.' is used.
%
\institute{Sun Yat-sen University, Zhuhai,   510275, China \\
\email{xiaofy5@mail2.sysu.edu.cn}
\and
Tsinghua Shenzhen International Graduate School, Tsinghua University, Shenzhen 518055, China\\
\email{chunminghe19990224@gmail.com} \and
Beijing Institute of Technology, Beijing, 100086, China \and
School of Computer Science and Technology, Ocean University of China, Qingdao, 266000, China}
% \end{comment}
%******************
\maketitle              % typeset the header of the contribution
\renewcommand{\thefootnote}{\fnsymbol{footnote}} %将脚注符号设置为fnsymbol类型，即特殊符号表示
\footnotetext[2]{This work was supported by the National Science Foundation of China under grant 62201538 and Natural Science Foundation of Shandong Province under grant ZR2022QF006.}
\renewcommand{\thefootnote}{\arabic{footnote}}

\begin{abstract}
% 150--250 words.
Concealed object segmentation (COS) is a challenging task that involves localizing and segmenting those concealed objects that are visually blended with their surrounding environments. Despite achieving remarkable success, existing COS segmenters still struggle to achieve complete segmentation results in extremely concealed scenarios. In this paper, we propose a Hierarchical Coherence Modeling (HCM) segmenter for COS, aiming to address this incomplete segmentation limitation. In specific, HCM promotes feature coherence by leveraging the intra-stage coherence and cross-stage coherence modules, exploring feature correlations at both the single-stage and contextual levels. 
Additionally, 
% HCM employs an object homogeneity loss to enhance internal consistency within the concealed object. Furthermore, 
we introduce the reversible re-calibration decoder to detect previously undetected parts in low-confidence regions, resulting in further enhancing segmentation performance. Extensive experiments conducted on three COS tasks, including camouflaged object detection, polyp image segmentation, and transparent object detection, demonstrate the promising results achieved by the proposed HCM segmenter.
% Abundant experiments comprehensively demonstrate that the proposed HCM achieves promising results in three COS tasks, including camouflaged object detection, polyp image segmentation, and transparent object detection.
% generating more complete segmentation maps for concealed object segmentation.

\end{abstract}

%% narrow the gap between equations and sentences
\setlength{\abovedisplayskip}{2pt}
\setlength{\belowdisplayskip}{2pt}

\section{Introduction}
% 页数规划：Intro:; Related Work: 到第三页为止; Methodology: 到六页半之前; Experiments: 剩下的三页半; Conclusions: .
% 总体10页+2页。切记10页不要抄，但是剩下的2页引满了，都是兄弟，都引一引。

% COS领域有一个挑战：1、由于内在一致性挑战，网络难以得到完整的分割，因此我们提出了分层一致性建模网络（特征聚类+特征约束——一致性约束；多个stage，分层，这是因为网络在深层具备深厚的语义信息，一致性约束可以促使网络把具备语义级相似性的特征聚合起来，而网络在浅层则更关于颜色、光照等底层信息，一个具备一致性约束能力的网络则可以进一步地对提取到的特征做像素级的微调。）
Concealed object segmentation (COS) is a challenging task with the purpose of localizing and segmenting those objects visually blended in their surrounding scenarios~\cite{fan2020camouflaged,he2023weaklysupervised}. COS is a general task encompassing various applications, such as camouflaged object detection (COD)~\cite{fan2020camouflaged}, polyp image segmentation (PIS)~\cite{fan2020pranet}, and transparent object detection (TOD)~\cite{mei2020don}. 

\begin{figure}[t]
	\centering
	\begin{subfigure}{0.19\textwidth}
		\centering
		\includegraphics[width=\textwidth]{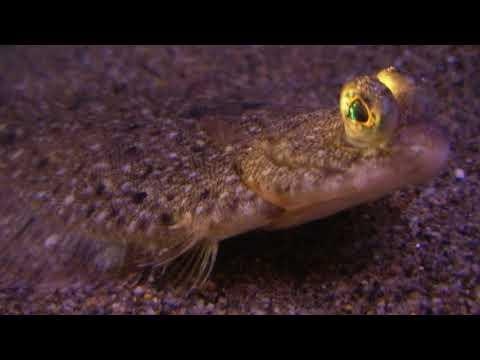}\vspace{-1pt}
	\end{subfigure}
	\hfill
	\begin{subfigure}{0.19\textwidth}  
		\centering 
		\includegraphics[width=\textwidth]{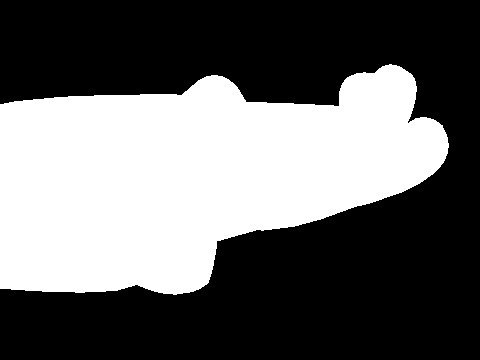}\vspace{-1pt}
	\end{subfigure}
	\hfill
	\begin{subfigure}{0.19\textwidth}  
		\centering 
		\includegraphics[width=\textwidth]{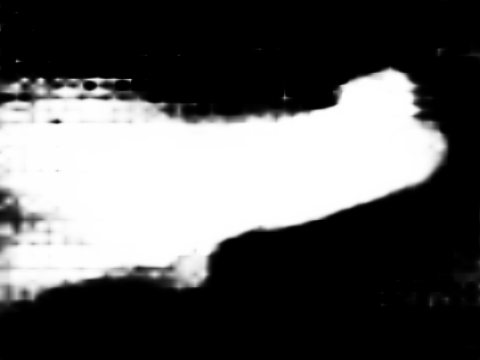}\vspace{-1pt}
	\end{subfigure}
    \hfill
	\begin{subfigure}{0.19\textwidth}  
		\centering 
		\includegraphics[width=\textwidth]{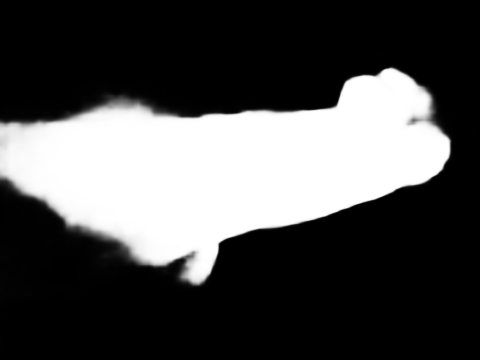}\vspace{-1pt}
	\end{subfigure}
	\hfill
	\begin{subfigure}{0.19\textwidth}   
		\centering 
		\includegraphics[width=\textwidth]{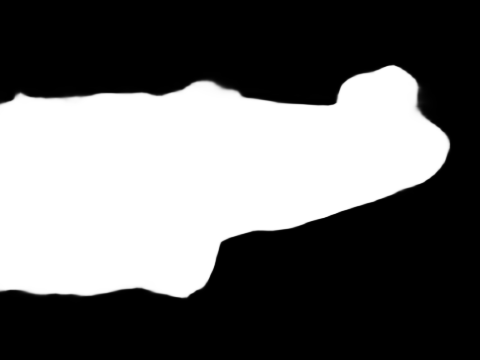}\vspace{-1pt}
	\end{subfigure} \\ \vspace{1mm}
	\begin{subfigure}{0.19\textwidth}
		\centering
		\includegraphics[width=\textwidth]{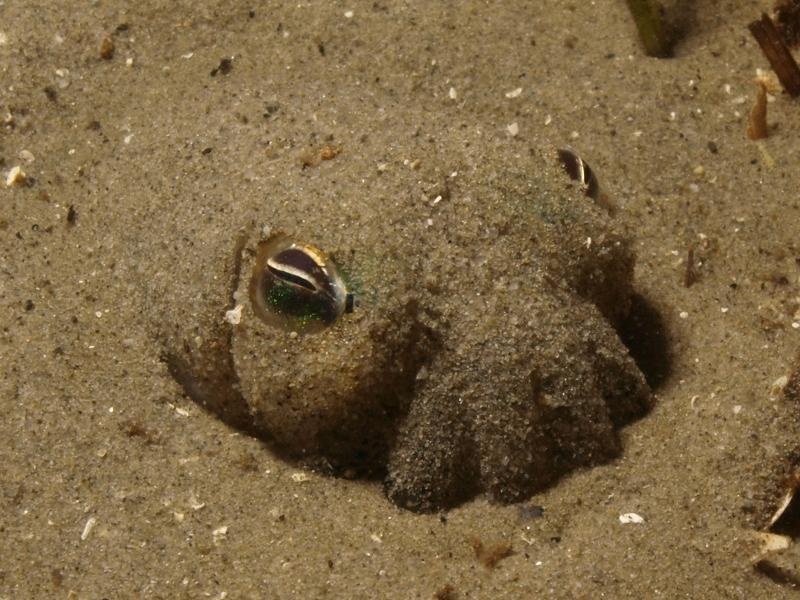}\vspace{-5pt}
		\caption{\footnotesize Origin}
	\end{subfigure}
	\hfill
	\begin{subfigure}{0.19\textwidth}  
		\centering 
		\includegraphics[width=\textwidth]{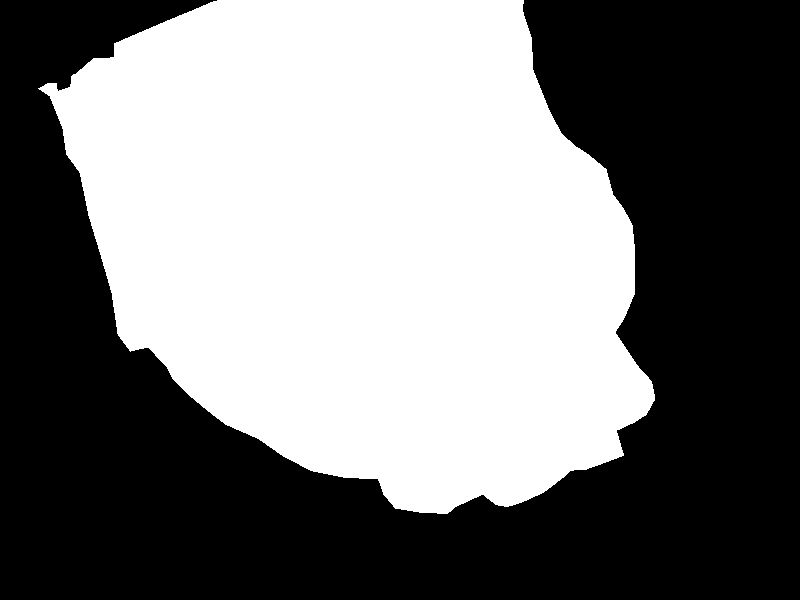}\vspace{-5pt}
		\caption{\footnotesize GT}
	\end{subfigure}
	\hfill
	\begin{subfigure}{0.19\textwidth}  
		\centering 
		\includegraphics[width=\textwidth]{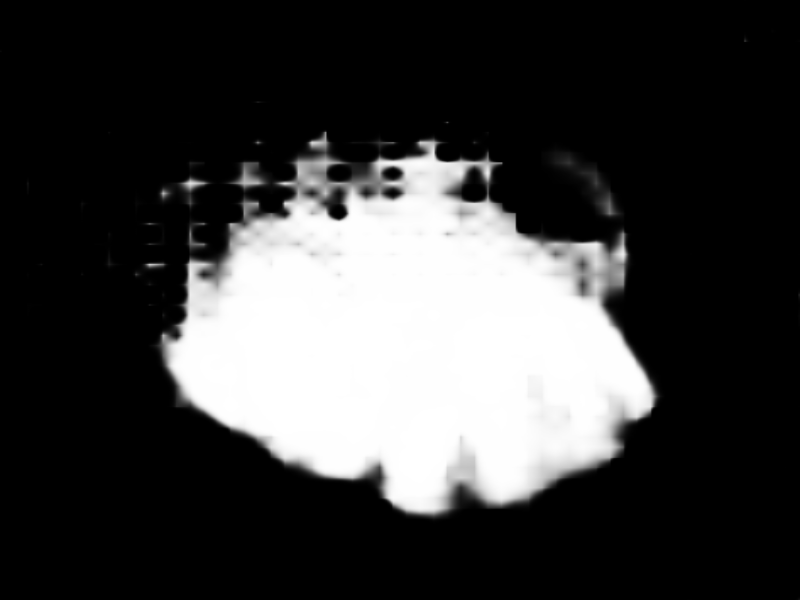}\vspace{-5pt}
		\caption{\footnotesize UGTR}
		%\label{fig:residual}
	\end{subfigure}
    \hfill
	\begin{subfigure}{0.19\textwidth}  
		\centering 
		\includegraphics[width=\textwidth]{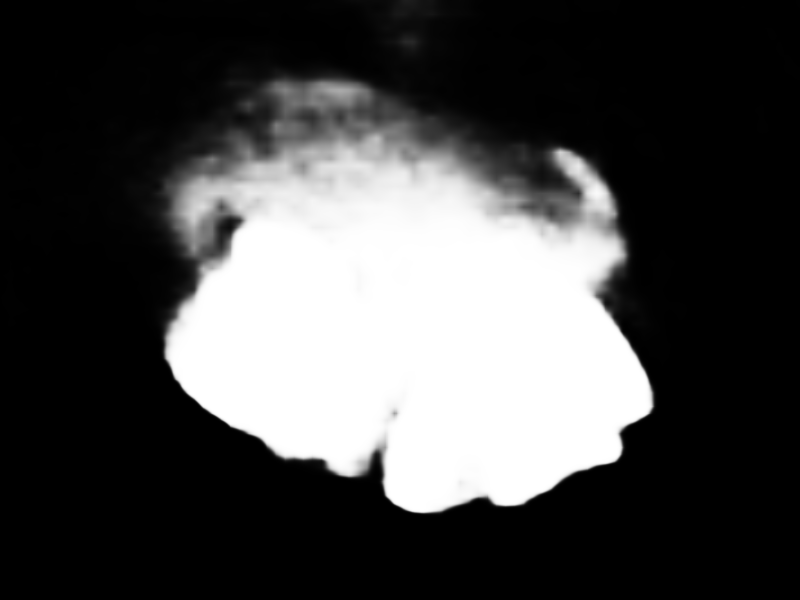}\vspace{-5pt}
        \caption{\footnotesize SegMaR}
	\end{subfigure}
	\hfill
	\begin{subfigure}{0.19\textwidth}   
		\centering 
		\includegraphics[width=\textwidth]{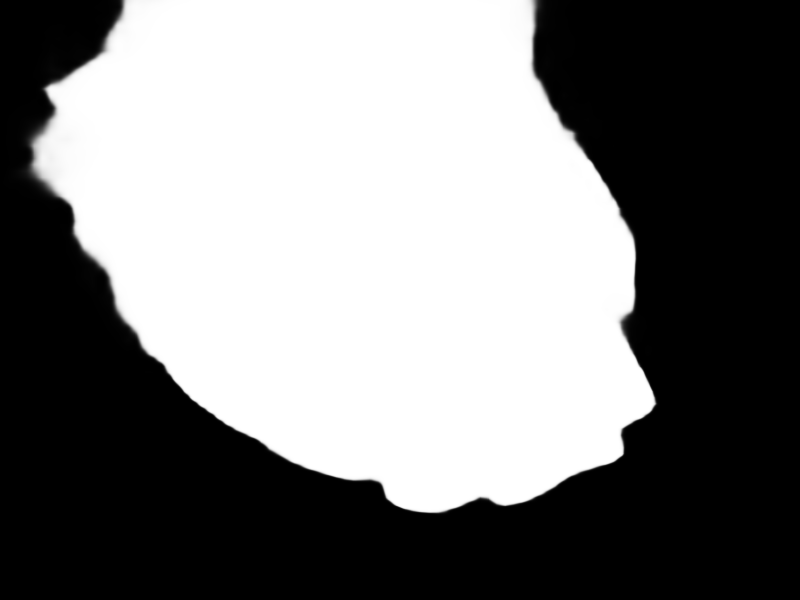}\vspace{-5pt}
		\caption{\footnotesize HCM} 
	\end{subfigure}\vspace{-3mm}
	\caption{ \small Results of UGTR~\cite{yang2021uncertainty}, SegMaR~\cite{jia2022segment}, and the proposed HCM. It is observed that our HCM can generate more accurate and complete results.
 %Our method exhibits better localization capacity and generates more complete results with clearer edges.
 }
	\label{fig:challenges}
	% \vspace{-4mm}
\end{figure}

COS poses significant challenges due to the intrinsic similarity between foreground objects and their corresponding scenarios, making it difficult to identify discriminative cues for complete and accurate foreground-background separation. To cope with this challenge, existing COS segmenters have employed various strategies, \textit{e.g.}, drawing inspiration from human vision~\cite{mei2021camouflaged,pang2022zoom,he2023strategic}, incorporating frequency clues~\cite{He2023Camouflaged}, and adopting joint modeling strategies across multiple tasks~\cite{zhai2021mutual,lu2021vector}. Despite their notable achievements, existing segmenters still struggle to achieve \textit{precise} results in some extremely concealed scenarios. As shown in Fig.~\ref{fig:challenges}, while UGTR~\cite{yang2021uncertainty} and  SegMaR~\cite{jia2022segment} manage to find the rough regions for the concealed objects, the prediction results are still incomplete.

To overcome this limitation, we propose a Hierarchical Coherence Modeling (HCM) segmenter for the COS task, which aims to generate more complete segmentation maps by promoting feature coherence. HCM incorporates two key components, namely, intra-stage coherence (ISC) and cross-stage coherence (CSC), to explore feature correlations at both single stage and contextual levels. 
% Furthermore, HCM utilizes an object homogeneity loss to reduce the variability of concealed internal features, thereby promoting internal consistency within the concealed object. 
Additionally, we develop the reversible re-calibration decoder (RRD) to detect previously undetected parts in those low-confidence regions and thus further improve segmentation performance.

Our contributions are summarized as follows:
\begin{itemize}
    \item We propose the Hierarchical Coherence Modeling (HCM) segmenter for the COS task. HCM encourages feature coherence and thus alleviating the incomplete segmentation problem.
    \item We introduce RRD to detect previously undetected parts in low-confidence regions, thus improving segmentation performance.
    \item The proposed HCM significantly outperforms the state-of-the-art methods on three COS tasks by a large margin, \textit{i.e.},
    camouflaged object detection, polyp image segmentation, and transparent object detection.
\end{itemize}

% COS is a challenging task for the lack of discriminative clues in concealed objects and has been investigated in various domains, such as camouflaged object detection (COD)~\cite{fan2020camouflaged}, polyp image segmentation (PIS)~\cite{fan2020pranet}, and transparent object detection (TOD)~\cite{mei2020don}.

\section{Related Works}
\noindent \textbf{Concealed Object Segmentation}. Traditional COS techniques heavily rely on manually designed feature extraction operators, which inherently suffer from limited feature extraction capacity and struggle to handle extremely complex scenarios. In contrast, learning-based approaches, facilitated by the rapid development of deep learning, have achieved remarkable success in this field. For instance, MGL~\cite{zhai2021mutual} introduces an auxiliary edge reconstruction task and constructs a mutual graph learning strategy to generate prediction maps with clear boundaries. Inspired by human vision principles~\cite{mei2021camouflaged}, PFNet leverages distraction mining techniques to achieve accurate concealed object segmentation. Recognizing the limitations of human vision, FEDER~\cite{He2023Camouflaged} introduces an adaptive decomposition approach to extract subtle yet discriminative clues that may have been overlooked. Unlike existing COS solvers, we propose HCM segmenter for the COS task, which encourages feature coherence and thus alleviates the incomplete segmentation problem. Additionally, we introduce the reversible re-calibration decoder (RRD) to detect previously undetected parts in low-confidence regions, further enhancing the segmentation performance. 

\begin{figure}[t]
	\centering
	\setlength{\abovecaptionskip}{-0.2cm}
	\begin{center}
		\includegraphics[width=0.9\linewidth]{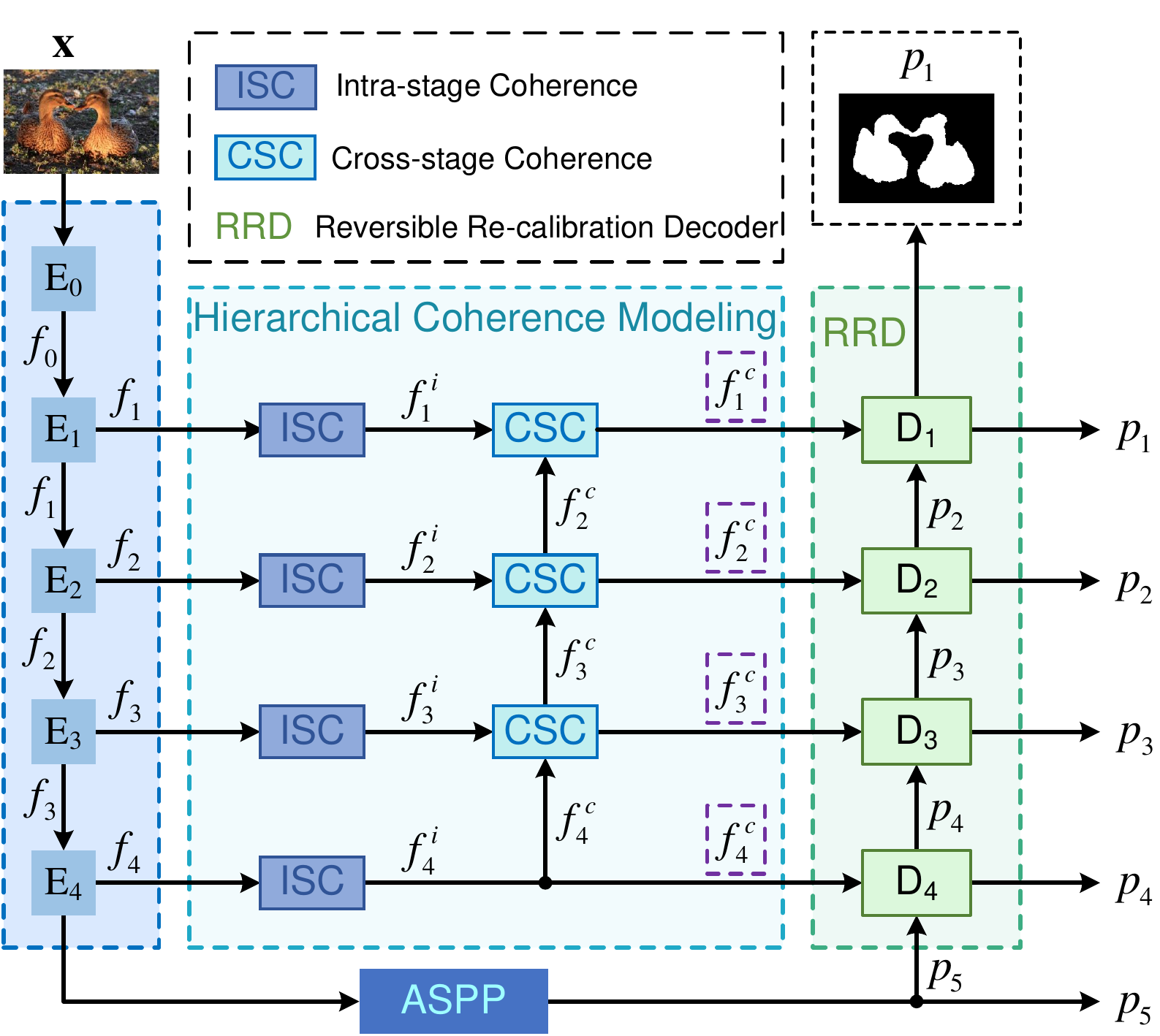}
	\end{center}
 \caption{Architecture of the proposed HCM. }
	\label{fig:Framework}
	% \vspace{-0.5cm}
\end{figure}
\begin{figure}[ht]
	\centering
	\setlength{\abovecaptionskip}{-0.2cm}
	\begin{center}
		\includegraphics[width=\linewidth]{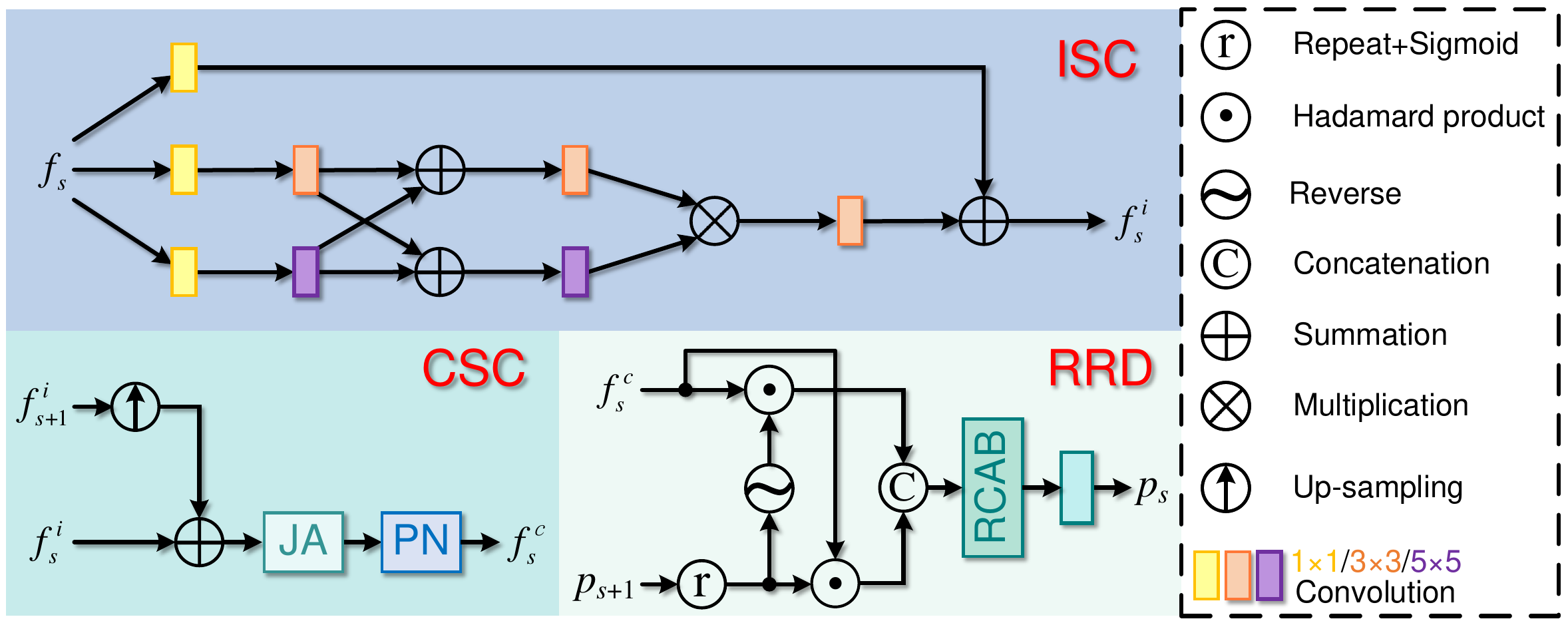}
	\end{center}
	\caption{Details of ISC, CSC, and RRD. }
	\label{fig:modules}
	% \vspace{-0.5cm}
\end{figure}
\section{Methodology}

\subsection{Concealed Feature Encoder}
Follow~\cite{fan2020camouflaged}, we employ ResNet50~\cite{he2016deep,he2023HQG} as the default backbone of the basic encoder $E$ for feature extraction. Given a concealed image $\mathbf{X}$, we obtain a series of feature maps $\{f_s\}_{s=0}^4$. Considering that $f_4$ has abundant semantic information, we further feed this feature map into an astrous spatial pyramid pooling (ASPP) module $A_s$~\cite{hu2021toward} to generate a coarse segmentation map $p_5$: $p_5=A_s(f_4)$.

\subsection{Hierarchical Coherence Modeling}
% 这里之所以这么命名，就是用的FEDER的写法
% COS领域有一个挑战：1、由于内在一致性挑战，网络难以得到完整的分割，因此我们提出了分层一致性建模网络（特征聚类+特征约束——一致性约束；多个stage，分层，这是因为网络在深层具备深厚的语义信息，一致性约束可以促使网络把具备语义级相似性的特征聚合起来，而网络在浅层则更关于颜色、光照等底层信息，一个具备一致性约束能力的网络则可以进一步地对提取到的特征做像素级的微调。）
Due to the inherent similarity between concealed objects and their surrounding contexts, obtaining accurate and complete segmentation results poses a significant challenge for the segmenter. To address this issue, we propose a hierarchical coherence modeling strategy that explicitly promotes feature coherence to facilitate more comprehensive predictions. Illustrated in Fig.~\ref{fig:modules}, this strategy consists of two parts: intra-stage coherence (ISC) and cross-stage coherence (CSC). These parts work in conjunction to enhance the coherence of features at different stages, promoting better segmentation outcomes.

\noindent \textbf{Intra-stage coherence}. ISC aims to discover feature correlations by fusing multi-scale features with different receptive fields within a single stage. This allows the aggregated features to capture scale-invariant information. As illustrated in Fig.~\ref{fig:modules}, ISC comprises two primary branches with residual connections. Given the input feature $f_s$, we initially apply a $1\times1$ convolution to reduce the channel dimension. Subsequently, we process these features with two parallel convolutions using different kernel sizes, resulting in features $f_s^3$ and $f_s^5$:
\begin{equation}
    f_s^3=conv3(conv1(f_s)), f_s^5=conv5(conv1(f_s)),
\end{equation}
where $conv1, conv3, conv5$ denote $1\times1, 3\times3, 5\times5$ convolutions, respectively. We proceed by merging the features $f_s^3$ and $f_s^5$, which are obtained from the previous step. These merged features are then further processed using two parallel convolutions. Finally, we multiply the outputs of these convolutions in a residual connection structure to extract the scale-invariant information. This process yields the aggregated features $\{f_s^i\}_{s=1}^4$:
% resulting in aggregated features $\{f_s^i\}_{s=1}^4$:
% We then combine $f_s^3$ and $f_s^5$, process them with another two parallel convolutions, and multiply the outputs to excavate the scale-invariant information:
\begin{equation}
    f_s^i=conv1(f_s)+conv3(conv3(f_s^3\oplus f_s^5)\otimes conv5(f_s^3\oplus f_s^5)),
\end{equation}
where $\oplus$ and $\otimes$ denote pixel-level summation and multiplication. 
% We then integrate the three features and acquire the aggregated feature $\{f_s^i\}_{s=1}^4$:
% \begin{equation}
%     f_s^i= conv1(f_s)+conv3(),
% \end{equation}
This design facilitates the extraction of features at multiple scales and enhances the ability to capture diverse contextual information.

% ISC seeks feature correlations by combining the multi-scale feature with various reception fields in a single stage. By doing so, the aggregated features can capture scale-invariant information. As depicted in Fig~\ref{}, ISC consists of two main branches with residual connections. Given feature $f_s$, we first employ a $1\times1$ convolution to reduce the channel number and then processing those features with two parallel convolutions with distinct kernel sizes, thus getting $f_s^3$ and $f_s^5$

\noindent \textbf{Cross-stage coherence}. CSC explores the contextual feature correlations by selectively interacting cross-stage information with joint attention $JA(\bigcdot)$, comprising spatial attention and channel attention~\cite{hu2021blind,deng2022pcgan}. Additionally, we employ position normalization $PN(\bigcdot)$ to highlight the contextual similarity and eliminate discrepancy interference information across different stages, getting $\{f_s^c\}_{s=1}^4$:
\begin{equation}
    f_s^c=PN(JA(f_s^i,up(f_{s+1}^i))),
\end{equation}
where $up(\bigcdot)$ denotes up-sampling operator.

% \noindent \textbf{Object homogeneity loss}. Aside from excavating the coherence information, we further propose an object homogeneity loss to decrease the variance of the concealed internal features, encouraging the internal consistency of the concealed object, which is defined as follows:
% \begin{equation}
%     L_{OH}= \sum_{s=1}^4 \|\mathbf{Y}\otimes f_s^c-P_s^o\|^2,
% \end{equation}
% where $\mathbf{Y}$ is the ground truth, which is resized to cater to the spatial resolution of the feature $f_s^c$. $P_s^o$ denotes the feature-level prototypes of the concealed objects.

% Given that the network captures deep semantic information at the deeper stages and focuses on underlying information, such as color and illumination, at the shallow stages, we incorporate the proposed hierarchical coherence modeling strategy across all four stages of our HCM segmenter. This strategy enables the HCM segmenter to effectively aggregate features with semantic-level similarity while also fine-tuning the extracted features at the pixel level. By combining these capabilities, our HCM segmenter achieves simultaneous enhancement of semantic-level feature aggregation and pixel-level refinement.

\subsection{Reversible Re-calibration Decoder}
Due to the complexity of concealed object scenes, segmenters produce prediction maps that contain low-confidence and ambiguous regions. To tackle this challenge, we propose a novel module called Reversible Re-calibration Decoder (RRD). As shown in Fig.~\ref{fig:modules}, RRD leverages both the previous decoder's prediction map as prior information and reverses the prediction map to extract cues from the low-confidence regions. This allows the segmenter to effectively detect previously undetected parts in these regions, leading to improved segmentation performance. Consequently, the prediction map $\{p_{s}\}_{s=1}^4$ is defined as follows:
% The prediction maps of segmenters trained with sparse annotations often contain low-confidence ambiguous regions due to the complex scenes of concealed objects and the limited discriminative capacity of the segmenter. To address this issue, we propose a reversible re-calibration decoder (RRD). RRD simultaneously utilizes the previous decoder's prediction map as prior information and reverses the prediction map to excavate cues from the low-confidence regions. By doing so, the segmenter can more effectively detect the undetected parts in these regions and thus improve the segmentation performance. Therefore, the prediction map  is defined as follows:
% given the feature map $f_s^l$ and the former prediction map $p_{s+1}$, 
\begin{equation}
        p_s = conv3\left(RCAB\left(cat\left(f_k^s \odot S\left(rp\left(p_{s+1}\right)\right),f_k^s \odot rv\left(S\left(rp\left(p_{s+1}\right)\right)\right)\right)\right)\right),
\end{equation}    
where $rp(\bigcdot)$, $S(\bigcdot)$, $rv(\bigcdot)$, $\odot$, and $conv3(\bigcdot)$ denote repeat, Sigmoid, reverse (element-wise subtraction with 1), Hadamard product, and $3\times 3$ convolution. $RCAB(\bigcdot)$ is the residual channel attention block~\cite{xu2023dm,ju2022ivf} and we employ this block to emphasize the noteworthy information.

\subsection{Loss Functions}
% Our HCM is supervised in a multiscale manner, \textit{i.e.}, $\{p_s\}_{s=1}^5$. 
We follow the practice in~\cite{fan2020camouflaged,He2023Camouflaged} and employ the weighted binary cross-entropy loss  $L_{BCE}^w$~\cite{xu2022multi} and weighted intersection-over-union loss $L_{IoU}^w$~\cite{rahman2016optimizing} to supervised our HCM with the ground truth $\mathbf{Y}$ in a multiscale manner
% , \textit{i.e.}, $\{p_s\}_{s=1}^5$, 
,which is defined as follows:
% two types of supervisory information:  sparse annotations and generated pseudo-labels. To ensure consistency between the segmentation results and the sparse annotations $\mathbf{Y}_i$, we use the partial cross-entropy loss $L_{pce}$, as employed in SCOD~\cite{he2022weakly}.
% Following SCOD~\cite{he2022weakly}, we employ the partial cross-entropy loss $L_{pce}$ to bind the consistency between the segmentation results and the sparse annotations. 
% For the pseudo-labels $\hat{\mathbf{Y}}_i$, we adopt the cross-entropy loss $L_{ce}$ and the intersection-over-union loss $L_{IoU}$, following the approach in~\cite{fan2021concealed}. Consequently, the overall loss function is defined as follows: 
\begin{equation}
    L = \sum_{s=1}^5 \frac{1}{2^{s-1}} \left(L_{BCE}^w\left(p_s, \mathbf{Y}\right)+L_{BCE}^w\left(p_s, \mathbf{Y}\right)\right)+L_{OH}.
\end{equation}

\section{Experiments}
\noindent \textbf{Implementation details}. Our HCM is implemented on two RTX3090TI GPUs and is optimized by Adam with the momentum terms $\left(0.9, 0.999\right)$. Following~\cite{fan2020camouflaged}, our encoder is initialized with the model pre-trained on ImageNet. During the training phase, the batch size is set to 32. The learning rate is initialized to 0.0001 and is divided by 10 every 80 epochs. The images are resized as $352\times352$.

\begin{table}[t]
\centering
\caption{Quantitative comparisons of our method and other 9 ResNet50-based SOTAs on COD. The best two results are in {\color[HTML]{FF0000} \textbf{red}} and {\color[HTML]{00B0F0} \textbf{blue}} fonts.}
\resizebox{\columnwidth}{!}{
\setlength{\tabcolsep}{1mm}
\begin{tabular}{l|c|cccc|cccc|cccc}
\toprule[1.5pt]
\multicolumn{1}{l|}{} & \multicolumn{1}{c|}{} & \multicolumn{4}{c|}{\textit{CAMO} (250 images)} & \multicolumn{4}{c|}{\textit{COD10K} (2,026 images)} & \multicolumn{4}{c}{\textit{NC4K} (4,121 images)} \\ \cline{3-14}
\multicolumn{1}{l|}{\multirow{-2}{*}{Methods}} & \multicolumn{1}{c|}{\multirow{-2}{*}{Pub.}} & {\cellcolor{gray!40}$M$~$\downarrow$} &{\cellcolor{gray!40}$F_\beta$~$\uparrow$} &{\cellcolor{gray!40}$E_\phi$~$\uparrow$} & \multicolumn{1}{c|}{\cellcolor{gray!40}$S_\alpha$~$\uparrow$}& {\cellcolor{gray!40}$M$~$\downarrow$} &{\cellcolor{gray!40}$F_\beta$~$\uparrow$} &{\cellcolor{gray!40}$E_\phi$~$\uparrow$} & \multicolumn{1}{c|}{\cellcolor{gray!40}$S_\alpha$~$\uparrow$}& {\cellcolor{gray!40}$M$~$\downarrow$} &{\cellcolor{gray!40}$F_\beta$~$\uparrow$} &{\cellcolor{gray!40}$E_\phi$~$\uparrow$} & \multicolumn{1}{c}{\cellcolor{gray!40}$S_\alpha$~$\uparrow$}\\ \midrule
% \multicolumn{14}{c}{ResNet-50-based COD} \\ \midrule
PFANet~\cite{zhao2019pyramid} & CVPR19  & 0.132  & 0.607  & 0.701  & 0.695 & 0.074 & 0.478 & 0.729 & 0.716 & 0.095 & 0.634 & 0.760  & 0.752 \\
CPD~\cite{wu2019cascaded}                                           & CVPR19      & 0.113                                 & 0.675                                 & 0.723                                 & 0.716                                 & 0.053                                 & 0.578                                 & 0.776                                 & 0.750                                 & 0.072                                 & 0.719                                 & 0.808                                 & 0.787                                 \\
EGNet~\cite{zhao2019egnet}                                        & ICCV19                                             & 0.109                                 & 0.667                                 & 0.800                                 & 0.732                                 & 0.061                                 & 0.526                                 & 0.810                                 & 0.736                                 & 0.075                                 & 0.671                                 & 0.841                                 & 0.777                                 \\
SINet~\cite{fan2020camouflaged}                                         & CVPR20                                             & 0.092                                 & 0.712                                 & 0.804                                 & 0.745                                 & 0.043                                 & 0.667                                 & 0.864                                 & 0.776                                 & 0.058                                 & 0.768                                 & 0.871                                 & 0.808                                 \\
PFNet~\cite{mei2021camouflaged}                                         & CVPR21                                             & 0.085                                 & 0.751                                 & 0.841                                 & 0.782                                 & 0.040                                 & 0.676                                 & 0.877                                 & 0.800                                 & 0.053                                 & 0.779                                 & 0.887                                 & 0.829                                 \\
MGL-R~\cite{zhai2021mutual}                                         & CVPR21                                             & 0.088                                 & 0.738                                 & 0.812                                 & 0.775                                 & \color[HTML]{00B0F0} \textbf{0.035}                                 & 0.680                                 & 0.851                                 & 0.814                                 & 0.053                                 & 0.778                                 & 0.867                                 & 0.833                                 \\
MGL-S~\cite{zhai2021mutual}                                         & CVPR21                                            & 0.089                                 & 0.733                                 & 0.806                                 & 0.772                                 & 0.037                                 & 0.666                                 & 0.844                                 & 0.811                                 & 0.055                                 & 0.771                                 & 0.862                                 & 0.829                                 \\
LSR~\cite{lv2021simultaneously}                                           & CVPR21                                             & 0.080                                 & 0.756                                 & 0.838                                 & 0.787                                 & 0.037                                 & \color[HTML]{00B0F0} \textbf{0.699}                                & 0.880                                 & 0.804                                 & 0.048                                 & \color[HTML]{00B0F0} \textbf{0.802}                                 & \color[HTML]{00B0F0} \textbf{0.890}                                & 0.834                                 \\
UGTR~\cite{yang2021uncertainty}                                          & ICCV21                                             & 0.086                                 & 0.747                                 & 0.821                                 & 0.784                                 & 0.036                                 & 0.670                                 & 0.852                                 & \color[HTML]{00B0F0} \textbf{0.817}  & \color[HTML]{00B0F0} \textbf{0.052} & 0.778  & 0.874 & \color[HTML]{00B0F0} \textbf{0.839} \\
SegMaR~\cite{jia2022segment} & CVPR22 & \color[HTML]{00B0F0} \textbf{0.072} & \color[HTML]{00B0F0} \textbf{0.772} & \color[HTML]{00B0F0} \textbf{0.861} & \color[HTML]{00B0F0} \textbf{0.805} & \color[HTML]{00B0F0} \textbf{0.035} & \color[HTML]{00B0F0} \textbf{0.699} & \color[HTML]{00B0F0} \textbf{0.890} & 0.813 & \color[HTML]{00B0F0} \textbf{0.052}  & 0.767                                 & 0.885                                 & 0.835                                 \\
\rowcolor{c2!20}HCM (Ours) &\multicolumn{1}{c|}{---}& {\color[HTML]{FF0000} \textbf{0.070}} & {\color[HTML]{FF0000} \textbf{0.782}} & {\color[HTML]{FF0000} \textbf{0.873}} & {\color[HTML]{FF0000} \textbf{0.806}} & {\color[HTML]{FF0000} \textbf{0.032}} & {\color[HTML]{FF0000} \textbf{0.736}} & {\color[HTML]{FF0000} \textbf{0.902}} & {\color[HTML]{FF0000} \textbf{0.820}} & {\color[HTML]{FF0000} \textbf{0.046}} & {\color[HTML]{FF0000} \textbf{0.816}} & {\color[HTML]{FF0000} \textbf{0.900}} & {\color[HTML]{FF0000} \textbf{0.846}} \\ 
\bottomrule[1.5pt]
\end{tabular}}
\label{table:CODQuanti}
% \vspace{-0.2cm}
\end{table}

\begin{figure}[t]
	\centering
	\setlength{\abovecaptionskip}{-0.2cm}
	\begin{center}
		\includegraphics[width=\linewidth]{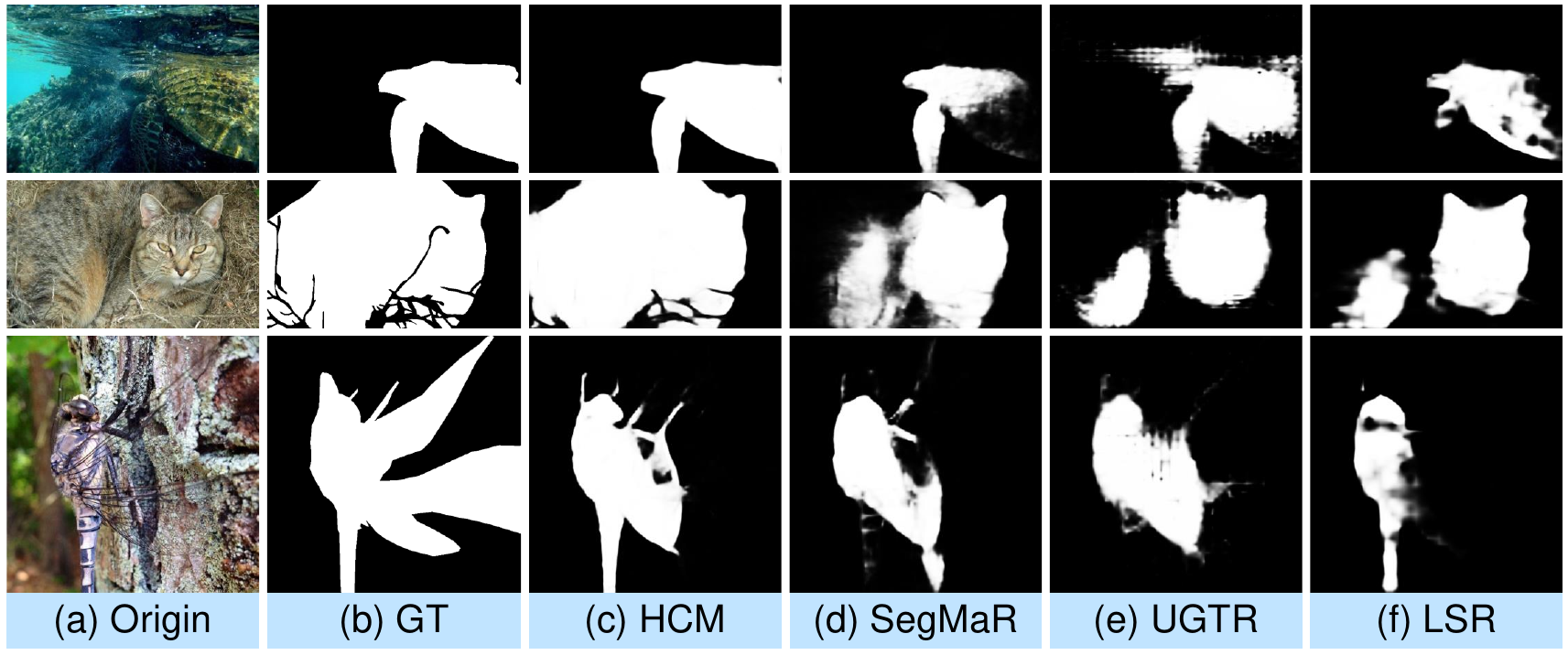}
	\end{center}
 \caption{Qualitative analysis on the COD task.}
	\label{fig:CODQuanli}
	% \vspace{-0.5cm}
\end{figure}

\begin{table}[t]
	\setlength{\abovecaptionskip}{0cm} 
	\setlength{\belowcaptionskip}{-0.2cm}
	\centering
    \caption{Quantitative comparisons on two benchmarks in polyp image segmentation. The best two results are in {\color[HTML]{FF0000} \textbf{red}} and {\color[HTML]{00B0F0} \textbf{blue}} fonts.} \label{table:PISQuanti}
	\resizebox{\columnwidth}{!}{
		\setlength{\tabcolsep}{0.8mm}
	\begin{tabular}{l|cccccc|cccccc}
		\toprule[1.5pt]
		\multirow{2}{*}{Methods}                                                                                                               & \multicolumn{6}{c|}{\textit{CVC-ColonDB} (380 images)}                                                                                                                & \multicolumn{6}{c}{\textit{Kvasir} (100 images)}                                                                                                                 \\ \cline{2-13} 
		& \multicolumn{1}{c}{\cellcolor{gray!40}mDice~$\uparrow$} & \multicolumn{1}{c}{\cellcolor{gray!40}mIoU~$\uparrow$} & {\cellcolor{gray!40}$M$~$\downarrow$}                                  & {\cellcolor{gray!40}$F_\beta^w$~$\uparrow$}                               & {\cellcolor{gray!40}$E_\phi^{max}$~$\uparrow$}                               & \multicolumn{1}{c|}{\cellcolor{gray!40}$S_\alpha$~$\uparrow$}& \multicolumn{1}{c}{\cellcolor{gray!40}mDice~$\uparrow$} & \multicolumn{1}{c}{\cellcolor{gray!40}mIoU~$\uparrow$} & {\cellcolor{gray!40}$M$~$\downarrow$}                                  & {\cellcolor{gray!40}$F_\beta^w$~$\uparrow$}                               & {\cellcolor{gray!40}$E_\phi^{max}$~$\uparrow$}                               & \multicolumn{1}{c}{\cellcolor{gray!40}$S_\alpha$~$\uparrow$} \\ \midrule
		U-Net~\cite{ronneberger2015u}              & 0.512                                 & 0.444                                 & 0.061                                 & 0.498                                 & 0.776                                 & 0.712                                 & 0.818                                 & 0.746                                 & 0.055                                 & 0.794                                 & 0.893                                 & 0.858                                 \\
		% U-Net++~\cite{zhou2018unet++}            & 0.483                                 & 0.410                                 & 0.064                                 & 0.467                                 & 0.760                                 & 0.691                                 & 0.821                                 & 0.743                                 & 0.048                                 & 0.808                                 & 0.910                                 & 0.862                                 \\
		Atten-UNet~\cite{oktay2018attention}         & 0.466                                 & 0.385                                 & 0.071                                 & 0.431                                 & 0.724                                 & 0.670                                 & 0.769                                 & 0.683                                 & 0.062                                 & 0.730                                 & 0.859                                 & 0.828                                 \\
		SFA~\cite{fang2019selective}                & 0.469                                 & 0.347                                 & 0.094                                 & 0.379                                 & 0.765                                 & 0.634                                 & 0.723                                 & 0.611                                 & 0.075                                 & 0.670                                 & 0.849                                 & 0.782                                 \\
		PraNet~\cite{fan2020pranet}             & 0.709                                 & 0.640                                 & 0.045                                 & 0.696                                 & 0.869                                 & 0.819                                 & 0.898                                 & 0.840                                 & 0.030                                 & 0.885                                 & 0.948                                 & 0.915                                 \\
		MSNet~\cite{zhao2021automatic}              & 0.755                                 & 0.678                                 & 0.041                                 & 0.737                                 & \color[HTML]{00B0F0} \textbf{0.883} & 0.836                                 & 0.907                                 & {\color[HTML]{00B0F0} \textbf{0.862}} & 0.028                                 & 0.893                                 & 0.944                                 & 0.922 \\
		TGANet~\cite{tomar2022tganet}             & 0.722                                 & 0.661                                 & 0.043                                 & 0.711                                 & 0.875                                 & 0.823                                 & 0.902                                 & 0.845                                 & 0.030                                 & 0.891                                 & 0.952                                 & 0.920                                 \\
		LADK~\cite{zhang2022lesion}               & \color[HTML]{00B0F0} \textbf{0.764} & 0.683                                 & 0.039                                 &  \color[HTML]{00B0F0} \textbf{0.739} & 0.862                                 & 0.834                                 & 0.905                                 & 0.852                                 & 0.028                                 & 0.887                                 & 0.947                                 & 0.918                                 \\
		$\text{M}^2\text{SNet}$~\cite{zhao2023m}             & 0.758                                 & \color[HTML]{00B0F0} \textbf{0.685} & {\color[HTML]{FF0000} \textbf{0.038}} & 0.737                                 & 0.869                                 & \color[HTML]{00B0F0} \textbf{0.842} & {\color[HTML]{FF0000} \textbf{0.912}} & 0.861                                 & {\color[HTML]{FF0000} \textbf{0.025}} & \color[HTML]{00B0F0} \textbf{0.901} & \color[HTML]{00B0F0} \textbf{0.953}& \color[HTML]{00B0F0} \textbf{0.922} \\
		\rowcolor{c2!20}HCM (Ours)  &\color[HTML]{FF0000} \textbf{0.775} &\color[HTML]{FF0000} \textbf{0.687} & \color[HTML]{FF0000}\textbf{0.038} &\color[HTML]{FF0000} \textbf{0.741} &\color[HTML]{FF0000} \textbf{0.885} &\color[HTML]{FF0000} \textbf{0.845} & \color[HTML]{00B0F0} \textbf{0.910} &\color[HTML]{FF0000} \textbf{0.868} &\color[HTML]{FF0000} \textbf{0.025} &\color[HTML]{FF0000} \textbf{0.903} & \color[HTML]{FF0000}\textbf{0.956} &\color[HTML]{FF0000} \textbf{0.924} \\
  \bottomrule[1.5pt]                           
	\end{tabular}}
	\vspace{-0.2cm}
\end{table}

\begin{figure}[t]
	\centering
	\setlength{\abovecaptionskip}{-0.2cm}
	\begin{center}
		\includegraphics[width=\linewidth]{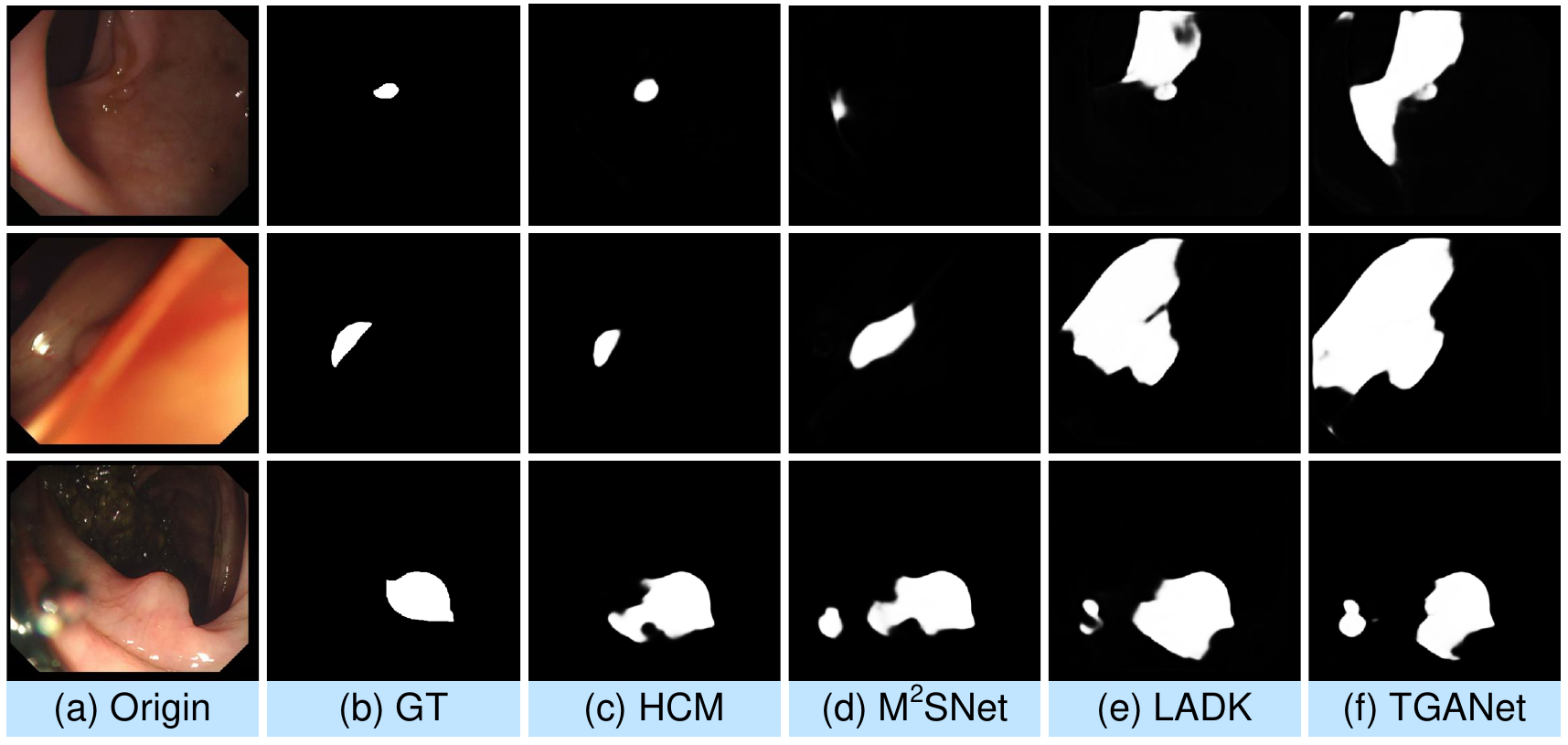}
	\end{center}
 \caption{Qualitative analysis on the PIS task.}
	\label{fig:PISQuanli}
	\vspace{-0.5cm}
\end{figure}

\begin{table*}[t]
	\setlength{\abovecaptionskip}{0cm} 
	\setlength{\belowcaptionskip}{-0.2cm}
	\centering
    \caption{ Quantitative comparisons on two benchmarks in transparent object detection. The best two results are in {\color[HTML]{FF0000} \textbf{red}} and {\color[HTML]{00B0F0} \textbf{blue}} fonts.} \label{table:TODQuanti}
	\resizebox{0.7\columnwidth}{!}{
		\setlength{\tabcolsep}{1mm}
	\begin{tabular}{l|cccc|cccc}\toprule[1.5pt]
		\multicolumn{1}{l|}{}                          & \multicolumn{4}{c|}{\textit{GDD} (936 images)}& \multicolumn{4}{c}{\textit{GSD} (810 images)}\\ \cline{2-9} 
		\multicolumn{1}{l|}{\multirow{-2}{*}{Methods}} & \cellcolor{gray!40}mIoU~$\uparrow$&\cellcolor{gray!40}$F_\beta^{max}$~$\uparrow$&\cellcolor{gray!40} $M$~$\downarrow$& \cellcolor{gray!40}BER~$\downarrow$&\cellcolor{gray!40}mIoU~$\uparrow$&\cellcolor{gray!40}$F_\beta^{max}$~$\uparrow$&\cellcolor{gray!40} $M$~$\downarrow$& \cellcolor{gray!40}BER~$\downarrow$ \\ \midrule
		PMD~\cite{lin2020progressive}& 0.870                                 & 0.930                                 & 0.067                                 & 6.17                                 & 0.817                                 & 0.890                                 & 0.061                                 & 6.74                \\
		GDNet~\cite{mei2020don}                                          & 0.876                                 & 0.937                                 & 0.063                                 & 5.62                                 & 0.790                                 & 0.869                                 & 0.069                                 & 7.72                                 \\
		% GateNet~\cite{zhao2020suppress}                                        & 0.817                                 & 0.931                                 & 0.073                                 & 8.84                                 & 0.689                                 & 0.898                                 & 0.073                                 & 10.12                                \\
		GlassNet~\cite{lin2021rich}                                       & 0.881                                 & 0.932                                 & 0.059                                 & 5.71                                 & 0.836                                 & 0.901                                 & 0.055                                 & 6.12                                 \\
		EBLNet~\cite{he2021enhanced}                                         & 0.870                                 & 0.922                                 & 0.064                                 & 6.08                                 & 0.817                                 & 0.878                                 & 0.059                                 & 6.75                                 \\
		CSNet~\cite{cheng2021highly}                                          & 0.773                                 & 0.876                                 & 0.135                                 & 11.33                                & 0.666                                 & 0.805                                 & 0.135                                 & 14.76                                \\
		PGNet~\cite{xie2022pyramid}                                          & 0.857                                 & 0.930                                 & 0.074                                 & 6.82                                 & 0.805                                 & 0.897                                 & 0.068                                 & 7.88                                 \\
		GDNet-B~\cite{mei2022large}                                        & 0.878                                 & 0.939                                 & 0.061                                 & 5.52                                 & 0.792                                 & 0.874                                 & 0.066                                 & 7.61                                 \\
		ESRNet~\cite{lin2022exploiting} & {\color[HTML]{00B0F0} \textbf{0.901}}  & {\color[HTML]{00B0F0} \textbf{0.942}} & {\color[HTML]{00B0F0} \textbf{0.046}} & {\color[HTML]{00B0F0} \textbf{4.46}} & {\color[HTML]{00B0F0} \textbf{0.854}} & {\color[HTML]{00B0F0} \textbf{0.911}} & {\color[HTML]{00B0F0} \textbf{0.046}} & {\color[HTML]{00B0F0} \textbf{5.74}} \\
		\rowcolor{c2!20}HCM (Ours) & {\color[HTML]{FF0000} \textbf{0.908}} & {\color[HTML]{FF0000} \textbf{0.946}} & {\color[HTML]{FF0000} \textbf{0.045}} & {\color[HTML]{FF0000} \textbf{4.42}} & {\color[HTML]{FF0000} \textbf{0.858}} & {\color[HTML]{FF0000} \textbf{0.922}} & {\color[HTML]{FF0000} \textbf{0.045}} & {\color[HTML]{FF0000} \textbf{5.52}} \\ \bottomrule[1.5pt]       
	\end{tabular}}
% \vspace{-0.2cm}
\end{table*}

\begin{figure}[t]
	\centering
	\setlength{\abovecaptionskip}{-0.2cm}
	\begin{center}
		\includegraphics[width=\linewidth]{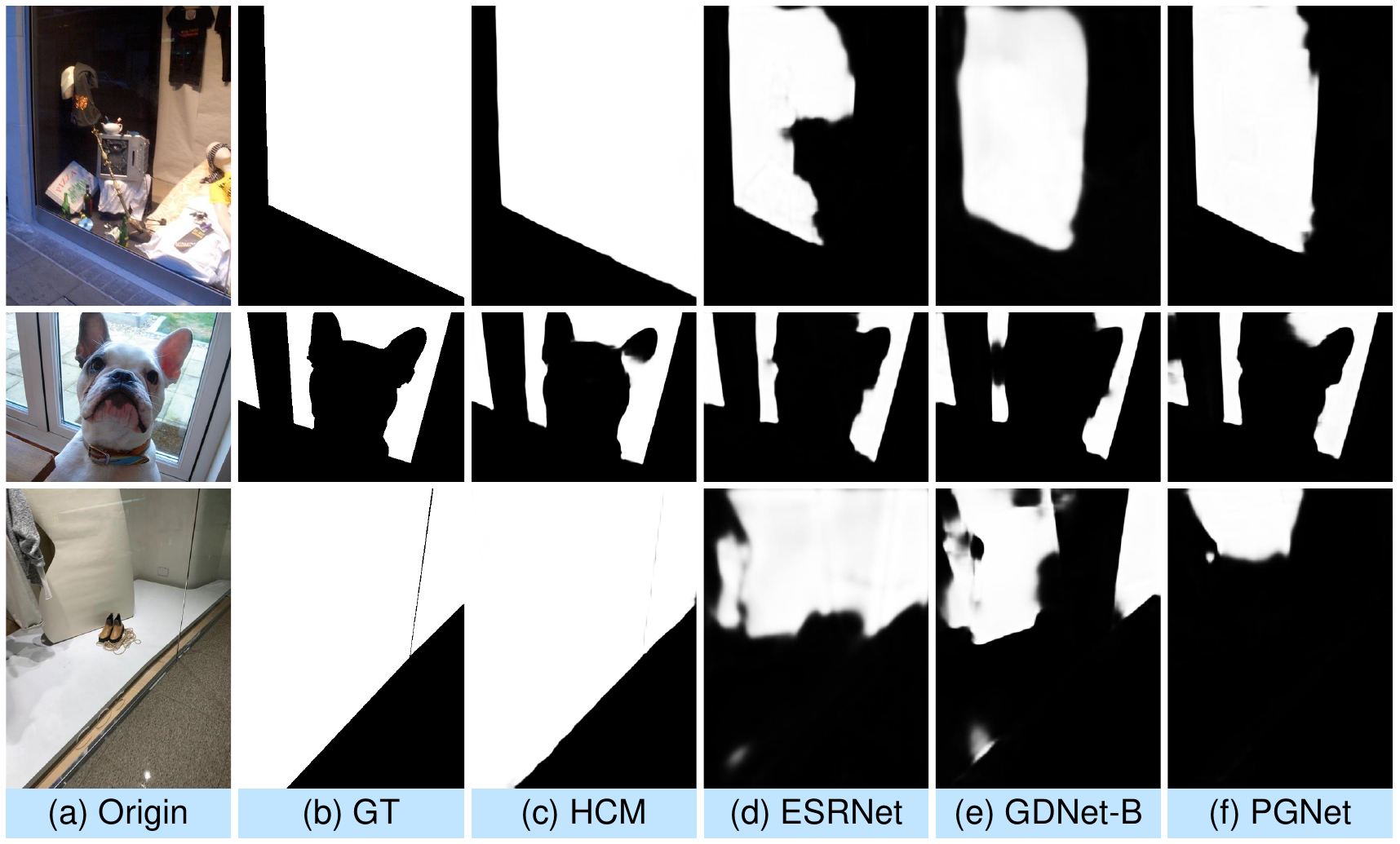}
	\end{center}
 \caption{Qualitative analysis on the TOD task.}
	\label{fig:TODQuanli}
	% \vspace{-0.5cm}
\end{figure}

% \subsection{Comparative Evaluation}

\subsection{Camouflaged Object Segmentation}
\noindent \textbf{Datasets and metrics}. 
Following~\cite{jia2022segment}, three datasets are utilized for evaluation, including \textit{CAMO}, \textit{COD10K}, and \textit{NC4K}. \textit{CAMO} comprises 1,250 camouflaged images with 8 categories. \textit{COD10K} have 10 super-classes, containing 5,066 images. $NC4K$ is the largest testing set which contains 4,121 images. Same as existing methods~\cite{fan2020camouflaged,jia2022segment}, our training set comprises 1,000 images from \textit{CAMO} and 3,000 images from \textit{COD10K}, and our test set integrates the rest of images.

Following~\cite{fan2020camouflaged,he2019image}, we utilize four metrics, namely mean absolute error $M$, adaptive F-measure $F_\beta$, mean E-measure $E_\phi$, and structure measure $S_\alpha$. Smaller $M$ means better performance, yet this is reversed on $F_\beta$, $E_\phi$, and $S_\alpha$.

\noindent \textbf{Quantitative analysis.} We compare the proposed Hierarchical Coherence Model (HCM) with nine other state-of-the-art ResNet50-based segmenters on the COD task and present the segmentation results in Table~\ref{table:CODQuanti}. As shown in the table, our HCM outperforms all other segmenters and achieves the top ranking. It surpasses the second-best COD segmenter, SegMaR~\cite{jia2022segment}, by a margin of 3.6\%. This remarkable performance clearly showcases the superior capability of our HCM in enhancing feature coherence.

% We make a comparison with other 9 ResNet50-based state-of-the-art segmenters on the COD task and report the segmentation results in Table~\ref{table:CODQuanti}. As depicted in Table~\ref{table:CODQuanti}, our HCM achieves the leading place and surpasses the second-best COD segmenter (SegMaR~\cite{jia2022segment}) over $3.6\%$. This fully demonstrates the superiority of our HCM in the enhancement of feature coherence.

\noindent \textbf{Quanliltative analysis.} As shown in Fig.~\ref{fig:CODQuanli}, our HCM demonstrates the capability to generate more complete and comprehensive segmentation results and reduces those uncertainty regions. This improvement can be attributed to our novel hierarchical coherence modeling strategy, which enhances feature coherence within the segmenter. Furthermore, the proposed reversible re-calibration decoder plays a crucial role in enabling the segmenter to effectively identify previously undetected parts within these regions.

\subsection{Polyp image segmentation}
\noindent \textbf{Datasets and metrics}. In line with the methodology employed in \cite{zhao2023m}, we evaluate the segmentation performance on two widely-used benchmark datasets: \textit{CVC-ColonDB} and \textit{Kvasir}. 900 images from \textit{Kvasir} make up the training set and the testing set comprises the remaining images. Additionally, Consistent with \cite{zhao2023m}, we adopt six metrics for quantitative evaluation: namely mean dice (mDice), mean IoU (mIoU), $M$, weighted F-measure ($F_\beta^w$), max E-measure ($E_\phi^{max}$), and $S_\alpha$. For mDice and mIoU, higher values indicate better performance, whereas for the remaining four metrics, higher values indicate poorer performance.

\noindent \textbf{Quantitative analysis.} Table~\ref{table:PISQuanti} presents the quantitative comparisons on two benchmarks in polyp image segmentation. As exhibited in Table~\ref{table:PISQuanti}, our proposed HCM outperforms the second-best techniques in $0.6\%$. This improvement can be attributed to the introduction of our novel modules, namely the Inter-Scale Coherence (ISC) and Contextual Scale Coherence (CSC) modules, which enable the exploration of feature correlations at both single-stage and contextual levels.

\noindent \textbf{Quanliltative analysis.} As shown in Fig~\ref{fig:PISQuanli}, our method can capture polyp more accurately because the proposed HCM method encourages feature coherence, which enables finer discrimination of the gap between foreground and background, leading to improved accuracy in polyp segmentation.

\subsection{Transparent object detection}
\noindent \textbf{Datasets and metrics}. 
In accordance with the experimental setup in \cite{lin2022exploiting}, we conduct our evaluations on two datasets: \textit{GDD} and \textit{GSD}. To assess the segmentation results, we employ four widely-used metrics: mean intersection over union (mIoU), maximum F-measure ($F_\beta^{max}$), $M$, and balance error rate (BER). The training set consists of 2,980 images from \textit{GDD} and 3,202 images from \textit{GSD}, while the remaining images are allocated to the testing set. It is important to note that a smaller value for $M$ or BER, or a higher value for IoU and $F_\beta^{max}$ indicates superior segmentation performance.

% Following~\cite{lin2022exploiting}, we adopt two datasets, namely \textit{GDD} and \textit{GSD}, and evaluate the segmentation results with four commonly-used metrics, including mIoU, max F-measure ($F_\beta^{max}$), $M$, and balance error rate (BER). The training set comprises 2,980 images from \textit{GDD} and 3,202 images from \textit{GSD}. The testing set contains the rest images. Notice that a smaller value on $M$, BER, or a higher value on IoU, $F_\beta^{max}$ indicates better performance.

\noindent \textbf{Quantitative analysis}. Table~\ref{table:TODQuanti} demonstrates the superior performance of our HCM in transparent object detection (TOD). Our method outperforms the second-best TOD solver, ESRNet, by $1.5\%$. This substantial improvement further validates the effectiveness and advancement of our proposed HCM segmenter in addressing the challenges of the TOD task.

\noindent \textbf{Quanliltative analysis}. As depicted in Figure~\ref{fig:TODQuanli}, our HCM segmenter achieves more accurate and complete segmentation results compared to the other methods. In contrast, the comparison methods often produce incomplete segments or exhibit blurred parts. These visual comparisons provide compelling evidence of the superiority of our method in addressing the challenges of incomplete segmentation with low-confidence regions.

\subsection{Ablation Study and Further Analysis}
% The proposed HCM segmenter includes two main components, the hierarchical coherence modeling 
We conduct ablation studies about our HCM on \textit{COD10K} of the COD task.

% {\cellcolor{gray!40}$M$~$\downarrow$} &{\cellcolor{gray!40}$F_\beta$~$\uparrow$} &{\cellcolor{gray!40}$E_\phi$~$\uparrow$} & \multicolumn{1}{c|}{\cellcolor{gray!40}$S_\alpha$~$\uparrow$}

\begin{table}[t]
\begin{subtable}{.57\textwidth}
	\centering
    \resizebox{1\columnwidth}{!}{
    \setlength{\tabcolsep}{0.6mm}
    \begin{tabular}{c|ccc}
    \toprule[1.5pt]
Metrics & w/o HCM component & w/o RRD & Ours  \\ \midrule
$M$~$\downarrow$    & 0.035             & 0.033   & \textbf{0.032} \\
$F_\beta$~$\uparrow$ & 0.702             & 0.725   & \textbf{0.736} \\
$E_\phi$~$\uparrow$ & 0.866             & 0.893   & \textbf{0.902} \\
$S_\alpha$~$\uparrow$ & 0.803             & 0.815   & \textbf{0.820} \\ \bottomrule[1.5pt]
\end{tabular}}\vspace{-5pt}
\caption{\fontsize{7.2pt}{\baselineskip}\selectfont Break down ablations of HCM.}\label{table:Breakdown}
\end{subtable}
\begin{subtable}{.412\textwidth}
\centering
    \resizebox{1\columnwidth}{!}{
    \setlength{\tabcolsep}{0.6mm}
    \begin{tabular}{c|ccc}
\toprule[1.5pt]
     Metrics & w/o ISC & w/o CSC  & Ours           \\ \midrule
$M$~$\downarrow$    & 0.033   & 0.033       & \textbf{0.032} \\
$F_\beta$~$\uparrow$ & 0.725   & 0.727         & \textbf{0.736} \\
$E_\phi$~$\uparrow$ & 0.895   & 0.888          & \textbf{0.902} \\
$S_\alpha$~$\uparrow$ & 0.812   & 0.815        & \textbf{0.820}    \\ \bottomrule[1.5pt]
\end{tabular}}\vspace{-5pt}
\caption{\fontsize{7.2pt}{\baselineskip}\selectfont Effect of the HCM strategy.}\label{table:EffectHCM}
    \end{subtable}
\caption{\small Ablation study on \textit{COD10K} of the COD task, where ``w/o'' denotes without.
% in the IVF task on the $\textit{M}^3\textit{FD}$ dataset. 
The best results are marked in \textbf{bold}.
}
		\label{table:Ablation} 
  % \vspace{-4mm}
\end{table}

\noindent \textbf{Break down ablations of HCM}. As demonstrated in Table~\ref{table:Breakdown}, when examining the individual components of our HCM, namely the HCM component or RRD, the performance of HCM decreases. This observation highlights the superiority of our proposed components in contributing to the overall performance of the proposed segmenter.

\noindent \textbf{Effect of the hierarchical coherence modeling strategy}. We conducted additional experiments to validate the effectiveness of each component in our proposed hierarchical coherence modeling strategy. As shown in Table~\ref{table:EffectHCM}, our results demonstrate the superiority of the ISC and CSC. These components work in synergy to form a powerful hierarchical coherence modeling strategy.

% \noindent \textbf{Performance of RRD in different stages}.

% \noindent \textbf{Performance on extreme scenarios}.

\section{Future Work and Conclusions}
In our future work, we will consider segmenting concealed objects in those degraded scenarios\cite{he2023degradation}, such as the low-light scenario and the underwater condition\cite{he2023reti}.

In this paper, we present a novel segmenter called HCM for COS with the aim of addressing the existing limitation of incomplete segmentation. The HCM method focuses on promoting feature coherence by utilizing both intra-stage coherence and cross-stage coherence modules, which explore feature correlations at both the single-stage and contextual levels. Moreover, we introduce the reversible re-calibration decoder to identify previously undetected parts in regions with low-confidence, thereby further improving the segmentation performance. The effectiveness of the proposed HCM segmenter is demonstrated through extensive experiments conducted on three different COS tasks: camouflaged object detection, polyp image segmentation, and transparent object detection. The results obtained from these experiments show promising outcomes, affirming the efficacy of the HCM approach.

%
% ---- Bibliography ----
%
% BibTeX users should specify bibliography style 'splncs04'.
% References will then be sorted and formatted in the correct style.
%
% \bibliographystyle{splncs04}
% \bibliography{mybibliography}
%

\clearpage
\bibliographystyle{splncs04}
\bibliography{HCM}
\end{document}